\documentclass[10pt,twocolumn,letterpaper,usenames,dvipsnames]{article}

\usepackage{wacv}
\usepackage{times}
\usepackage{epsfig}
\usepackage{graphicx}
\usepackage{amsmath}
\usepackage{amssymb}
\usepackage{subcaption}
\usepackage{booktabs}
\usepackage{subcaption}
\usepackage[final]{pdfpages}

\usepackage{xcolor,colortbl}

\definecolor{Gray}{gray}{0.85}
\newcolumntype{g}{>{\columncolor{Gray}}c}

\definecolor{floor}{RGB}{0,128,0}
\definecolor{furniture}{RGB}{0,128,128}
\definecolor{structure}{RGB}{0,0,255}
\definecolor{props}{RGB}{200,200,0}

\definecolor{oil_bottle}{RGB}{0,0,255}
\definecolor{tissue_box}{RGB}{255,0,0}
\definecolor{blue_funnel}{RGB}{153,153,0}
\definecolor{drill}{RGB}{153,200,200}
\definecolor{cracker_box}{RGB}{102,0,52}
\definecolor{tomato_soup}{RGB}{0,102,102}
\definecolor{spam}{RGB}{0,0,0}

\usepackage[pagebackref=true,breaklinks=true,letterpaper=true,colorlinks,bookmarks=false]{hyperref}

\wacvfinalcopy 

\def\wacvPaperID{62} 

\ifwacvfinal\fi

\begin{document}

\title{Casting Geometric Constraints in Semantic Segmentation\\ as Semi-Supervised Learning}


\author{Sinisa Stekovic$^1$\\
	\and
	Friedrich Fraundorfer$^1$\\
	\and
	Vincent Lepetit$^{2,1}$\\
	\and
	$^1$Institute for Computer Graphics and Vision, Graz University of Technology, Graz, Austria \\
	$^2$Universit\'e Paris-Est, \'Ecole des Ponts ParisTech, Paris, France \\
	{\tt\small \{sinisa.stekovic, fraundorfer, lepetit\}@icg.tugraz.at}
}

\maketitle
\ifwacvfinal\thispagestyle{empty}\fi

\begin{abstract}

We propose a simple yet effective method to learn to segment new indoor scenes from video frames: State-of-the-art methods trained on one dataset, even as large as the SUNRGB-D dataset, can perform poorly when applied to images that are not part of the dataset, because of the dataset bias, a common phenomenon in computer vision. To make semantic segmentation more useful in practice, one can exploit geometric constraints. Our main contribution is to show that these  constraints can be cast conveniently as semi-supervised terms, which enforce the fact  that the same class should be predicted for the projections of the same 3D location in different images. This is interesting as we can exploit general existing techniques developed for semi-supervised learning to efficiently incorporate the constraints. We show that this approach can efficiently and accurately learn to segment target sequences of ScanNet and our own target sequences using only annotations from SUNRGB-D, and geometric relations between the video frames of target sequences.

\end{abstract}



\section{Introduction}

\begin{figure}
  \begin{center}
  \begin{subfigure}[b]{.49\linewidth}
  \includegraphics[width=1.\linewidth]{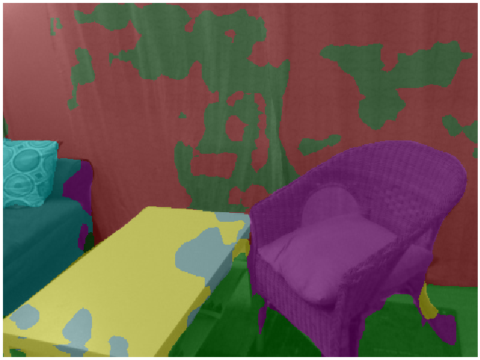}
  \caption{}
  \label{fig:intro_sb}
  \end{subfigure}
  \begin{subfigure}[b]{.49\linewidth}
  \includegraphics[width=1.\linewidth]{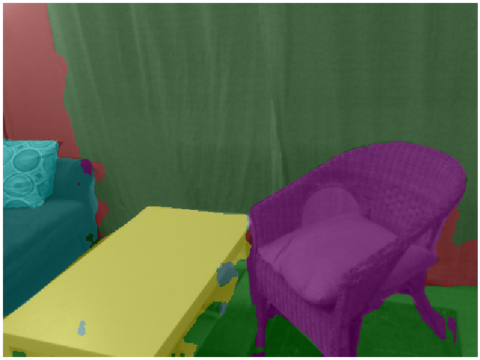}
  \caption{}
   \label{fig:intro_our}
  \end{subfigure}
  \end{center}
  \vspace{-0.5cm}
  \caption{(a) Even the state-of-the-art method DeepLabV3+ trained with training data from SUNRGB-D makes many mistakes when segmenting an image outside the SUNRGB-D dataset. (b) After exploiting geometric constraints on an unlabeled sequence of the new scene, our semi-supervised S4-Net approach predicts much better segmentations. }
  \label{fig:one}
\end{figure}


Semantic segmentation  of images provides  high-level understanding of  a scene,
useful for  many applications  such as robotics  and augmented  reality.  Recent
approaches can perform very well~\cite{Long2015,HeGDG17,zhao2017pyramid, deeplabv3plus2018}. 


In practice, however, it is difficult to generalize from existing datasets to new scenes. In other words, it is a challenging task to obtain good segmentation of images that do not belong to the training datasets. To demonstrate this, we trained a state-of-the-art segmentation method DeepLabV3+~\cite{deeplabv3plus2018} on the SUNRGB-D dataset~\cite{song2015sun}, which is made of more than $5000$ training images of indoor scenes. Fig.~\ref{fig:intro_sb} shows the segmentation we obtain when we attempt to segment a new image, which does not belong to the dataset. The performance is clearly poor, showing that the  SUNRGB-D dataset was not sufficient to generalize to this image, despite the size of the training dataset. 

To make semantic segmentation more practical and to break this dataset bias, one can exploit geometric constraints~\cite{ma2017multi, pham2019real, mccormac2017semanticfusion, grinvald2019volumetric}, in addition to readily available training data such as the SUNRGB-D dataset. We introduce an efficient formalization of this approach, which relies on the observation that geometric constraints can be introduced as standard terms from the semi-supervised learning literature. This  results in an elegant, simple, and powerful method that can learn to segment new environments from video frames,  which  makes it very useful for applications such as robotics and augmented reality.

More exactly, we adapt a general technique for  semi-supervised learning that consists of adding constraints on pairs of unlabeled training samples that are close to each other in the feature space,  to enforce  the fact  that such  two samples  should belong  to the  same category~\cite{LaineA17,   TarvainenV17,  athiwaratkun2018}. This is very close to what we want to do when enforcing geometric constraints for semantic segmentation: Pairs of unlabeled  pixels that correspond to the same physical 3D point would be labeled with the same category. 
In practice, to obtain the geometric information needed to enforce the constraints, we can rely on the measurements from depth sensors or train a network to predict depth maps as well, using recent techniques for monocular image reconstruction.

In contrast to previous methods exploiting geometric constraints for semantic segmentation, our method introduces several novelties. Comparing to \cite{ma2017multi}, our approach applies geometric constraints to completely unlabeled scenes. Furthermore, when compared to \cite{pham2019real, mccormac2017semanticfusion, grinvald2019volumetric}, which use simple label fusion to segment given target sequence, our approach can generalize from  a representation of one target sequence from the target scene to segmenting unseen images of the target scene. We demonstrate this aspect further in the evaluation section.

In short, our contribution is to show that semi-supervised learning is a simple yet principled and powerful  way to exploit geometric constraints in learning semantic segmentation.  We demonstrate  this by learning to annotate sequences of the  ScanNet~\cite{Dai2017} dataset using only annotations from the SUNRGB-D dataset. We also demonstrate effectiveness of the proposed method through the semantic labeling of our own newly generated sequence  unrelated to SUNRGB-D and ScanNet.

In the rest of the paper, we discuss related work, describe our approach, and present its evaluation with quantitative and qualitative experiments together with an ablation study.

\section{Related Work}

In this  section, we discuss  related work on  the aspects  of semantic
segmentation,  domain adaptation, general semi-supervised  learning,  and also  recent methods  for
learning depth  prediction from  single images, as  they also  exploit geometric
constraints similar to our approach. Finally, we discuss similarities and differences 
with other works that also combine segmentation and geometry.

\subsection{Supervised Semantic Segmentation with Deep Networks}

The introduction of deep  learning made a  large impact on performance of  semantic
segmentation.    Fully   Convolutional  Networks~(FCNs)~\cite{Long2015}   allow
segmentation prediction for  input of arbitrary size. In  this setting, standard
image classification task networks~\cite{SimonyanZ15a, HeZRS16} can be used by transforming fully-connected layers into convolutional ones.  FCNs use deconvolutional layers that  learn the interpolation  for upsampling  process.  Other works  including SegNet~\cite{BadrinarayananK17} and U-Net~\cite{RonnebergerFB15} rely  on similar architectures.  Such  works have been applied   to   a    variety   of   segmentational   tasks~\cite{RonnebergerFB15,  ArmaganHRL17, MiliotoLS18}. 
  
Recent methods address the problem of utilizing global context information for semantic segmentation. PSPNet~\cite{zhao2017pyramid} proposes to capture global context information through a pyramid pooling module that combines features under four different pyramid scales. DeepLabV3+~\cite{deeplabv3plus2018} uses atrous convolutions to control response of feature maps and applies atrous spatial pyramid pooling for segmenting objects at multiple scales. In our experiments, we apply our approach to both DeepLabV3+ and PSPNet to demonstrate it generalizes to  different network architectures. In principle, any other architecture could be used instead.

\subsection{Semi-Supervised Learning with Deep Networks}

Availability of  ground truth
labels is often the main limitation of supervised methods in pratice.  In contrast,  semi-supervised learning is a general  approach aiming at
exploiting both  labeled and unlabeled or weakly labeled training data.  Some approaches  rely on
adversarial    networks    to    measure   the    quality    of    unlabeled
data~\cite{DaiYYCS17, SantosWZ17, Kumar17, Hung2018}. More in line with our work
are   the   popular   consistency-based   models~\cite{LaineA17,   TarvainenV17,
  athiwaratkun2018}.  These  methods enforce the  model output to  be consistent
under  small  input  perturbations.   As  explained  in~\cite{athiwaratkun2018},
consistency-based models  can be viewed  as a student-teacher model:  To measure
consistency of model $f$, or the student, its predictions are compared to
predictions of a teacher model $g$,  a different trained model, while at the
same time applying small input perturbations.

$\Pi\mathit{-model}$~\cite{LaineA17}   is   a  recent   method   using   a
consistency-based  model  where the  student  is  its own  teacher,  \emph{i.e.}
$f=g$.  It relies on a cross-entropy  loss term applied to labeled data
only and an additional term that  penalizes differences in predictions for small
perturbations of input data.  Our semi-supervised approach is closely related to
the $\Pi\mathit{-model}$  but relies on  geometric consistency instead  of enforcing  consistent predictions for different input perturbations.

As pixel-level annotations, required for semantic segmentation tasks, are typically very time consuming to obtain, weakly-supervised methods become very interesting options for further increasing the amount of training samples. One way of obtaining more training data is through image-level annotations or bounding boxes. \cite{papandreou2015weakly} demonstrates that a network trained with large number of such weakly-supervised samples in combination with small amount of samples with pixel-level annotations achieves comparable results to a fully supervised approach. Given image-level annotations rather than pixel-level annotations, \cite{wei2018revisiting} generates dense object localization maps which are then utilized in a weakly- or semi-supervised framework to learn semantic segmentation. Our geometric constraints can be seen as a form of weak supervision but instead of weak labels our approach relies only fon weak constraints enforcing consistent annotations for  3D points of the scene.

\subsection{Domain Adaptation For Semantic Segmentation}

Domain adaptation has been studied for the field of semantic segmentation. One can argue that overcoming the dataset bias is closely related to the field of domain adaptation. In the context of semantic segmentation, such approaches usually leverage the possibility of using inexaustive synthetic datasets for improving performance on real data~\cite{luo2019significance, michieli2019adversarial, sun2019not, chang2019all}. However, as further explained in~\cite{sun2019not}, due to large domain gap between real and synthetic images, such domain adaptation methods easily overfit to synthetic data and can fail to generalize to real images.

Very recently, in terms of domain adaptation approaches that rely on real data only, Kalluri~\etal~\cite{kalluri2018} proposed a unified segmentation model for different target domains that minimizes  supervised loss for labeled data of the target domains and exploits visual similarity between unlabeled data of the domains. Results indicate increase in performance for all of the target domains. However, such approach still requires labeled images for all of the target domains. 
 Here, we focus on adapting the source domain to a related target domain for which no labeling is available.

\subsection{Geometric Constraints and Label Propagation}

Geometry in semantic segmentation has already been considered for purpose of semantic mapping. 
\cite{ma2017multi} trains a CNN by propagating manual hand-labeled segmentations of frames to new frames by warping. In contrast, we do not need any manual annotations for the target sequences of the scene. SemanticFusion~\cite{mccormac2017semanticfusion} uses a pre-trained CNN together with ElasticFusion SLAM method~\cite{WhelanSGDL16}, and merges multiple segmentation predictions from different viewpoints. \cite{pham2019real, grinvald2019volumetric} rely on using a pre-trained CNN together with 3D reconstruction methods and improve accuracy over initial segmentations. However, these approaches are applied to a CNN with fixed parameters and rely on geometric constraints during inference time. In contrast, our method uses geometric constraints to improve single-view segmentation predictions for the target scene and afterwards requires only color information for segmenting unseen images of the scene. 

\subsection{Single-View Depth Estimation}
Because  of  view warping,  our  approach  is also  related  to  recent work  on
unsupervised single-view depth estimation.  Both Zhou~\etal~\cite{ZhouBSL17} and
Godard~\etal~\cite{Godard17}  proposed  an  unsupervised approach  for  learning
depth estimation from  video data. This is  done by learning to  predict a depth
map so  that a  view can be  warped into another  one.  This  research direction
became   quickly    popular,   and   has    been   extended   since    by   many
authors~\cite{Yin18, Mahjourian18, Wang18, Godard18}.

Our work is related to these methods as it  also introduces constraints between multiple views, by  using warping. We demonstrate that this type of constraints can be utilized for the task of semantic segmentation.


\section{Approach Overview}

\newcommand{\calS}{\mathcal{S}}
\newcommand{\calU}{\mathcal{U}}
\newcommand{\calW}{\mathcal{W}}
\newcommand{\calWP}{\mathcal{\WP}}
\newcommand{\WA}{W\!\!A}
\newcommand{\WP}{W\!\!P}
\newcommand{\NV}{N\!\!V}
\newcommand{\Warp}[2]{\underset{#1\rightarrow#2}{\text{Warp}}}
\newcommand{\Merge}[1]{\underset{#1}{\text{Merge}}}
\newcommand{\calN}{\mathcal{N}}
\newcommand{\CE}{{\text{CE}}}
\newcommand{\WCE}{{\text{WCE}}}

\begin{figure}[t]
	\centering
	\begin{subfigure}[t]{1.\linewidth}
	    \centering
	    \includegraphics[width=1.\linewidth]{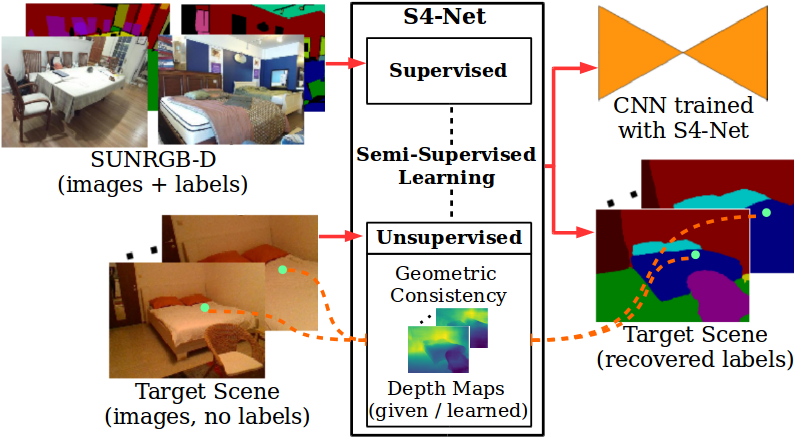}
    	\caption{}
        \label{fig:method}
	\end{subfigure}
    \hspace{2cm}
    \begin{subfigure}[t]{.95\linewidth}
    \centering
    \includegraphics[width=1.\linewidth]{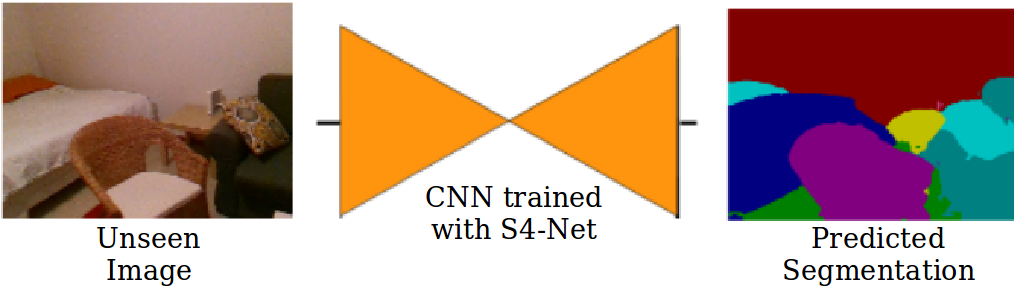}
    \caption{}
    \end{subfigure}
	\caption{Method overview. (a) Our S4-Net approach combines supervised data from SUNRGB-D and an image sequence from a target scene without any annotations. By exploiting geometric constraints of the target image sequence, we obtain a network with high performance for the target scene, and labels for the target sequence. (b) After being trained by S4-Net, the segmentation network can be applied to unseen images of the target scene with much better performance.}
	\label{fig:method_over}
\end{figure}

For the rest of the paper, we refer to our Semi-Supervised method for Semantic Segmentation as S4-Net. We assume that we are given a dataset of color images and their segmentations:
\[
\calS = \{e_i = (I_i, A_i)\}_i \> ,
\] 
where $I_i$ is a color image, and $A_i$ is the corresponding ground truth segmentation. In practice we use the SUNRGB-D dataset. Based on these annotations, we would like to train a segmentation model $f()$ for a new scene given a sequence of registered frames, for which no labels are known \emph{a priori}:
\[
\calU = \{e_j = (I_j, D_j, T_j)\}_j \> ,
\]
where $I_j$ is a color image, $D_j$ is the corresponding depth map, and $T_j$ the corresponding camera pose. As a direct result of S4-Net, we obtain automatic annotations for sequence $\calU$. Additionally, the output of S4-Net is a trained network $f()$. At test time, network $f()$ trained with S4-Net can then be used to predict correct segmentation for new images of the scene. We present the method overview in Fig.~\ref{fig:method_over}.


\subsection{Semi-Supervised Learning and Geometric Consistency}

We optimize the  parameters $\Theta$ of $f()$ by minimizing the semi-supervised loss term:
\begin{equation}
L = L_S + \lambda L_G \> ,
\label{eq:loss}
\end{equation}
where $L_S$ is a supervised loss term and $L_G$ is a term that exploits geometric constraints. In practice, we set the discount factor $\lambda$ for all experiments to the same value. $L_S$ is a standard term for supervised learning of semantic segmentation:
\begin{equation}
L_S = \sum_{e\in\calS} l_\WCE(f(I(e);\Theta), A(e)),
\end{equation}
where $l_\WCE$ is the weighted cross-entropy of segmentation prediction $f(I;\Theta)$ relative to manual annotation $A$. The class weights are calculated using median frequency balancing~\cite{eigen2015} to prevent overfitting to most common classes.

$L_G$ exploits geometric constraints to enforce consistency between predictions for images taken from different viewpoints:
%
\begin{equation}
L_G = \sum_{e\in\calU} l_\CE(f(I(e);\Theta), \Merge{e' \in \calN(e)}(\Warp{e'}{e}(f(I(e'); \Theta '))) \> ,
\end{equation}
where $\calN(e)$ is a subset of $\calU$ containing samples with a viewpoint that overlaps with the view point of $e$. $\Warp{e'}{e}(S)$ function warps segmentation $S$ from frame $e'$ to frame $e$. We give more details on this warp operation in Section~\ref{sec:warping}. $\Merge{e' \in \calN(e)}$ function merges given neighbouring views by first summing the pixelwise probabilities and then performing $argmax$ operation to obtain the final pixelwise labels.

We consider prediction $f(I(e'); \Theta ')$ as a teacher prediction and, similarly to the $\Pi\mathit{-model}$~\cite{LaineA17}, it is treated as a constant when calculating the update of the network parameters. Parameters $\Theta '$ are updated every $100$ iterations to equal parameters $\Theta$. We found that this step helps to further stabilize the learning process. $l_\CE$ is the standard cross-entropy loss function that compares the predicted segmentations. We found empirically that using weighted cross-entropy tends to converge to solutions with incorrect segmentations. 



  




\subsection{Segmentation Warping}
\label{sec:warping}

We base our warping function $\Warp{e'}{e}$ on the inverse warping method used in~\cite{ZhouBSL17}. For a 2D location $p$ in homogeneous coordinates of a target sample $e$, we find the corresponding location $p'$ of the source sample $e'$ using:
\begin{align}
p' = K T_{e \rightarrow e'} d K^{-1} p \> ,
\end{align}
where $K$ is the intrinsic matrix of the camera, $T_{e \rightarrow e'}$ is the relative transformation matrix between the target and the source samples, and $d$ is the predicted depth value at location $p$. Since $p'$ value lies in general between different image locations, we  use the differentiable  bilinear
interpolation from~\cite{JaderbergSZK15} to  compute the final projected value from the 4 neighbouring pixels. This transformation is applied to the segmentation probabilities predicted by the network.

\begin{figure*}
    \centering
    \begin{subfigure}[b]{0.24\linewidth}
     \centering
    \includegraphics[width=1.\linewidth]{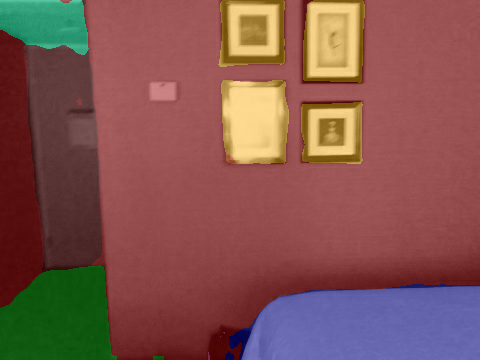}
    \subcaption{}
    \end{subfigure}%
    \hfill
    \begin{subfigure}[b]{0.24\linewidth}
     \centering
    \includegraphics[width=1.\linewidth]{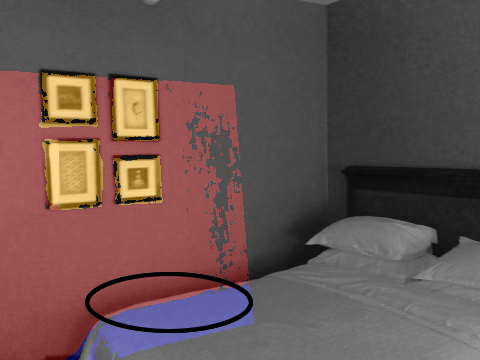}
    \subcaption{}
    \end{subfigure}%
    \hfill
    \begin{subfigure}[b]{0.24\linewidth}
     \centering
    \includegraphics[width=1.\linewidth]{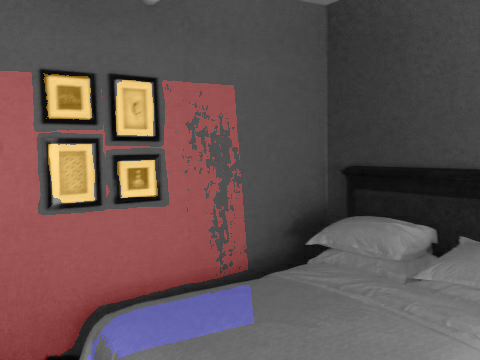}
    \subcaption{}
    \end{subfigure}%
    \hfill
    \begin{subfigure}[b]{0.24\linewidth}
     \centering
    \includegraphics[width=1.\linewidth]{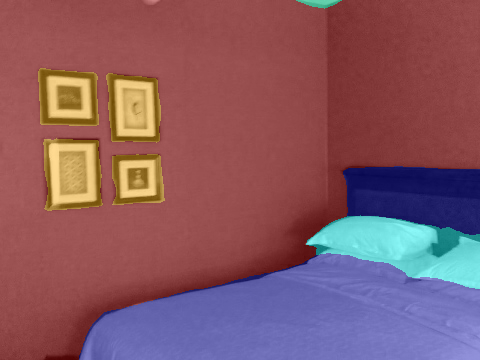}
    \subcaption{}
    \end{subfigure}%

    \caption{Process of warping source segmentation prediction in (a) to the corresponding target view. Encircled region in (b) demonstrates that warping in boundary regions of segmentation predictions can induce errors in the target view. In this case, the warped segmentation prediction falsely assigns wall labels to the edges of the bed region. Hence, we introduce a segmentation boundary mask to resolve this issue in (c). As direct result, S4-Net is able to recover quality segmentations in the affected boundary regions in (d).}
    \label{fig:seg_mask}
\end{figure*}

In practice,  not every  pixel in the  target sample has  a correspondent pixel in the
source sample.  This can happen as
depth information is not necessarily available for every pixel when using depth cameras, and
since some pixels  in the target sample  may not be visible in  the source sample,
because they  are occluded or, simply,  because they are  not in the field  of view of the
source sample. If the  difference between the depths is larger than a threshold value, this
means that the pixel is occluded and does not correspond to the same physical 3D
point. We simply ignore the pixels without correspondents in  the loss function. Additionally, we ignore the pixels that are located near the edges of the predicted segments: Segmentation predictions in these regions tend to be less reliable and, for such regions, insignificant errors in one view can easily induce significant errors in other views because of the different perspectives, as shown in Fig.~\ref{fig:seg_mask}


\subsection{S4-Net with Depth Prediction}
\label{sec:depth}

For sequences captured with an RGB camera, depth data is not available, and
we rely on predicted depths to enforce geometric constraints. If the supervised dataset $\calS$ also includes  ground truth depths, we can introduce additional loss terms to learn depth estimation:
\begin{equation}
L_D = L_{DS} + \lambda_D L_{DG} \> ,
\label{eq:loss}
\end{equation}
where $L_{DS}$ is a supervised depth loss term and $L_{DG}$ is a semi-supervised term that exploits geometric constraints in the depth domain. $\lambda_D$ is a weighting factor. $L_{DS}$ is the absolute difference loss term:
\begin{equation}
L_{DS} = \sum_{e\in\calS} |f_d(I(e);\Theta_d) - \hat{D}(e)| \> , 
\label{eq:loss}
\end{equation}
where $f_d(I;\Theta_d)$ is the depth prediction for network parameters $\Theta_d$ and $\hat{D}$ is the ground truth depth map. Term $L_{DG}$ corrects noisy depth predictions for the target scene $\calU$ through geometric constraints only:
\begin{equation}
L_{DG} = \sum_{e\in\calU} \sum_{e' \in \calN(e)} L_{INT}((I(e), \Warp{e'}{e}(I(e'))) \> ,
\end{equation}
where $L_{INT}$ is a  loss term comparing pixelwise intensities together with the structure similarity loss term from recent literature on monocular depth prediction~\cite{ZhouBSL17, Godard17, Yin18, Mahjourian18, Wang18, Godard18}. We apply this term only to the target image pixels where the predicted segmentations are consistent with each other and further away from segmentation borders: We found that such mask helps regularize depth predictions for occluded regions of the image. We explain this further in supplementary material. For enforcing geometric constraints on semantic segmentation with term $L_G$, we consider the depth prediction as a constant when calculating the update of the network parameters.


\subsection{Network Architecture}

We use DeepLabV3+~\cite{deeplabv3plus2018} or PSPNet~\cite{zhao2017pyramid} as network $f()$ in our experiments. In both cases, as the base network, we use ResNet-50~\cite{HeZRS16} pre-trained on the ImageNet classification task~\cite{deng2009imagenet}. However, S4-Net is not restricted to a specific type of architecture and could be applied to other architectures as well. When predicting depth maps, the encoder is shared between the depth network and the segmentation network. The depth decoder has the same architecture as the segmentation decoder, but they do not share any parameters. We show further details on network initialization and training procedure in supplementary material.






\section{Evaluation}
\label{sec:eval}

We evaluate S4-Net on the task of learning to predict semantic segmentation from color images for a target scene. The task for the network is to learn segmentations for a target scene without any knowledge about the ground truth labels for the scene. Hence, S4-Net requires an uncorrelated annotated dataset to obtain prior knowledge about the segmentation task that it needs to perform. Additionally, in order to learn accurate segmentations for the target scene, it utilizes frame sequences of that scene. By exploiting geometric constraints between the frames, S4-Net learns to predict segmentations across the target scene.  

\textbf{Datasets.} In all of our evaluations, we use the SUNRGB-D dataset~\cite{song2015sun}, consisting of $5285$ annotated RGB-D images for training, as the supervised dataset $\calS$ and perform mirroring on these samples to augment the supervised data. The SUNRGB-D dataset is a collection made of an original dataset and additional datasets previously published~\cite{SilbermanHKF12, janoch2013category, xiao2013sun3d}.  The images are manually segmented into $37$ object categories that are typical for an indoor scenario. The full list of object categories is given in supplementary materials. First, we evaluate S4-Net on scenes from the ScanNet dataset~\cite{Dai2017}. Second, we show that S4-Net is general by applying it to our own data. Finally, we show that enforcing geometric constraints through depth predictions can be used to learn quality segmentation predictions for the target scene.

\subsection{Evaluation on ScanNet}

\renewcommand{\arraystretch}{0.9} 
\newcommand{\sm}[1]{{\small #1}}

\begin{table*}[t]
 \centering
\begin{tabular}{@{}lcccccccrcccccc@{}}
  \toprule
  & \phantom{}&  \multicolumn{6}{c}{{\bf ``Scan 1'' (ScanNet)}} & \phantom{a}& \multicolumn{6}{c}{{\bf ``Scan 2'' (ScanNet)}}\\
  
    & \phantom{}&  \multicolumn{6}{c}{{(Unlabeled images during training)}} & \phantom{a}& \multicolumn{6}{c}{{(Excluded from training)}}\\
  
  && \sm{pix\_acc} & \sm{mean\_acc} & \sm{mIOU} & \sm{fwIOU} &&&&  \sm{pix\_acc} & \sm{mean\_acc} & \sm{mIOU} & \sm{fwIOU} &&   \\
  \midrule
  \multicolumn{6}{l}{\textbf{DeepLabV3+ network architecture}} &&&&&&&& \\ 
  
  Supervised baseline && $0.765$ & $0.634$ & $0.533$ & $0.692$ &&&& $0.772$ & $0.651$ & $0.544$ & $0.697$ && \\

  \textit{S4-Net} && $\textbf{0.803}$ & $\textbf{0.687}$ & $\textbf{0.59}$ & $\textbf{0.733}$ &&&& $\textbf{0.803}$ & $\textbf{0.691}$ & $\textbf{0.593}$ & $\textbf{0.732}$ && \\

  \textit{S4-Net with depth dred.} && $0.794$ & $0.679$ & $0.581$ & $0.724$ &&&& $0.797$ & $0.684$ & $0.586$ & $0.725$ && \\

\midrule
  \multicolumn{6}{l}{\textbf{PSPNet network architecture}} &&&&&&&& \\ 
  
  Supervised baseline && $0.727$ & $0.597$ & $0.486$ & $0.644$ &&&& $0.737$ & $0.61$ & $0.499$ & $0.654$ && \\

  \textit{S4-Net} && $\textbf{0.781}$ & $\textbf{0.648}$ & $\textbf{0.539}$ & $\textbf{0.701}$ &&&& $\textbf{0.78}$ & $\textbf{0.652}$ & $\textbf{0.546}$ & $\textbf{0.699}$ && \\


\bottomrule
\end{tabular}

\caption{Quantitative evaluation on the target scenes from ScanNet. We include results averaged over the target scenes used during experiments. The results for ``Scan 1'' show significant improvements for the images where we applied S4-Net. Furthermore, the segmentation accuracy for ``Scan 2'' indicates that our trained network brings similar improvements over the supervised baseline for the images that were not utilized by S4-Net during training. Our experiments also show that S4-Net can be applied to different segmentation networks with a significant gain in accuracy in comparison to the supervised baselines and the results with depth prediction show a comparable increase in performance over supervised baseline.}
\label{fig:scannet_quant}
\end{table*}

\begin{figure*}
    \centering
    \begin{subfigure}[b]{1.\linewidth}
     \centering
     \begin{minipage}{0.02\linewidth}
         \rotatebox[origin=t]{90}{Sup. baseline}
    \end{minipage}
    \begin{minipage}{0.23\linewidth}
        \includegraphics[width=\linewidth]{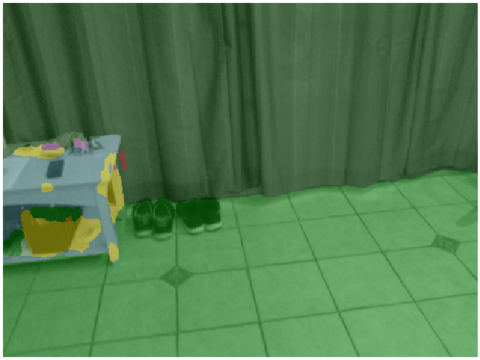}
    \end{minipage}
    \begin{minipage}{0.23\linewidth}
    \includegraphics[width=\linewidth]{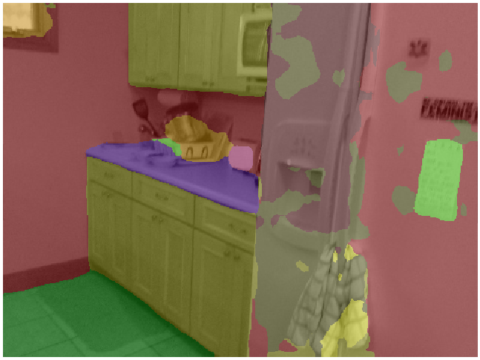}
    \end{minipage}
    \begin{minipage}{0.23\linewidth}
    \includegraphics[width=\linewidth]{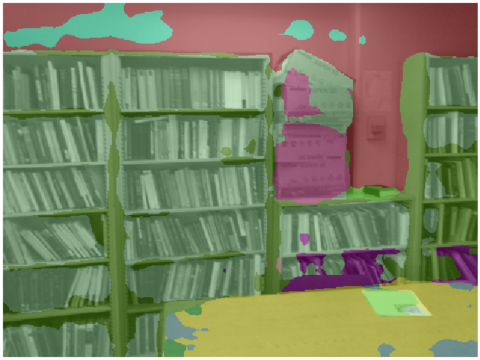}
    \end{minipage}
    \begin{minipage}{0.23\linewidth}
    \includegraphics[width=\linewidth]{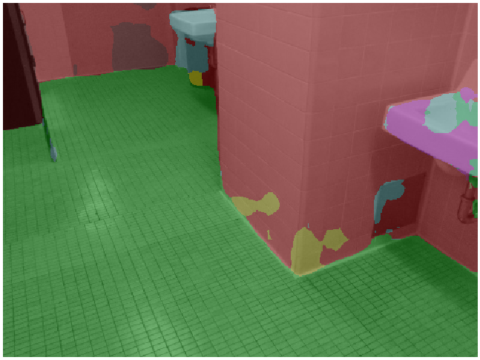}
    \end{minipage}
    \end{subfigure}
    \begin{subfigure}[b]{1.\linewidth}
     \centering
     \begin{minipage}{0.02\linewidth}
         \rotatebox[origin=t]{90}{\textit{S4-Net}}
    \end{minipage}
    \begin{minipage}{0.23\linewidth}
    \includegraphics[width=\linewidth]{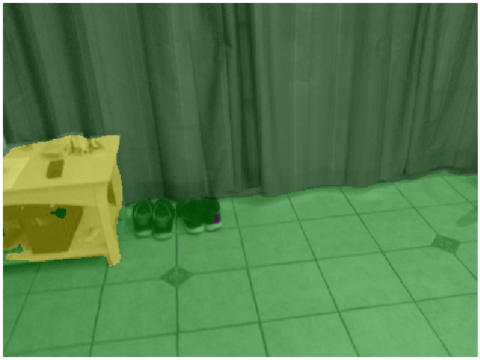}
    \end{minipage}
    \begin{minipage}{0.23\linewidth}
    \includegraphics[width=\linewidth]{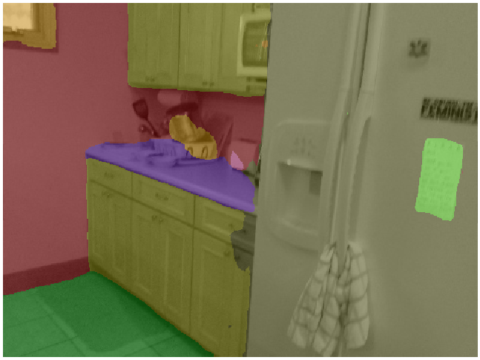}
    \end{minipage}
    \begin{minipage}{0.23\linewidth}
    \includegraphics[width=\linewidth]{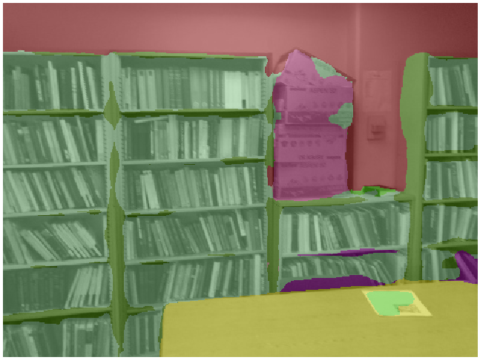}
    \end{minipage}
    \begin{minipage}{0.23\linewidth}
    \includegraphics[width=\linewidth]{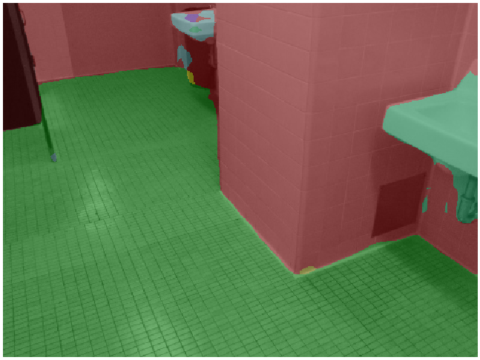}
    \end{minipage}
    \end{subfigure}
    \begin{subfigure}[b]{1.\linewidth}
     \centering
    \begin{minipage}{0.02\linewidth}
        \rotatebox[origin=t]{90}{\hspace{0.5cm} Man. annot.}
    \end{minipage}
    \begin{minipage}{0.23\linewidth}
    \includegraphics[width=\linewidth]{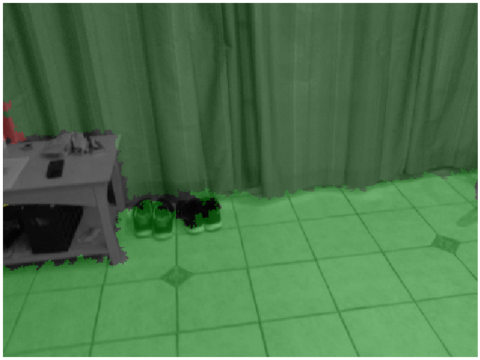}    \captionof{figure}{}
    \end{minipage}
    \begin{minipage}{0.23\linewidth}
    \includegraphics[width=\linewidth]{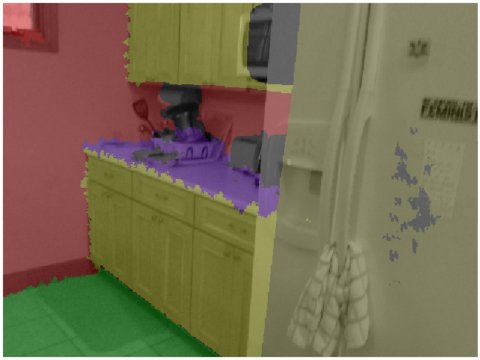}    \captionof{figure}{}
    \end{minipage}
    \begin{minipage}{0.23\linewidth}
    \includegraphics[width=\linewidth]{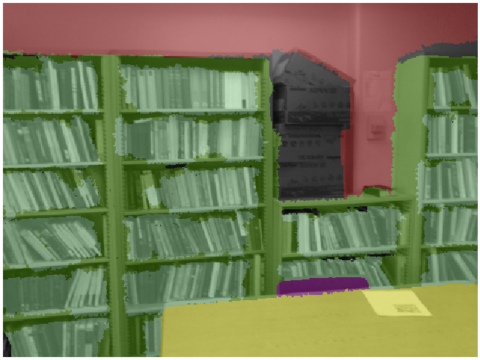}    \captionof{figure}{}
    \end{minipage}
    \begin{minipage}{0.23\linewidth}
    \includegraphics[width=\linewidth]{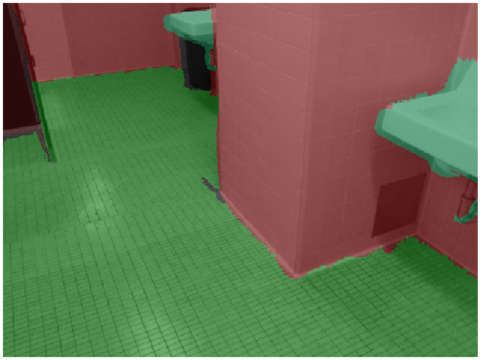}    \captionof{figure}{}
    \end{minipage}
    \end{subfigure}
    \caption{Qualitative results on unseen images from ``Scan 2'' of the target scenes from ScanNet, DeepLabV3+ network architecture. As S4-Net does not rely on manual annotations for the target scenes, it predicts segmentations that are sometimes more accurate than manual annotations. More specifically, it correctly segments the otherwise unlabelled table and box regions in (a) and (c), and in regions with wrong manual annotations it correctly predicts paper segments in (b) and (c).}
    \label{fig:scannet_qual}
\end{figure*}

As previously discussed, we use SUNRGB-D for the supervised training data only.  Therefore, for our first experimental setup, we evaluate S4-Net on $6$ scenes from the ScanNet dataset~\cite{Dai2017} to demonstrate the generalization aspect of S4-Net, as it is the scenario that motivates our work. These scenes represent different indoor scenarios, including apartment, hotel room, public bathroom, large kitchen space, lounge area, and study room. Even though the RGB-D sequences in the ScanNet dataset are annotated and can be mapped to our desired segmentation task, we utilize these annotations only to  validate our results. 

\textbf{Data Split.} Intentionally, we choose scenes from the ScanNet dataset which were scanned twice during the creation of the dataset. The first scan of each scene is utilized during training while the second scan is used for validation purposes only. We refer to these independently recorded scans as ``Scan 1'' and ``Scan 2''. During training,  we use the registered RGB-D sequence from ``Scan 1'' when applying geometric constraints. For evaluation, we additionally validate performance on ``Scan 2'' for which the camera follows a different pathway.


In this experiment, we show that our network, trained with geometric constraints from a target scene of ScanNet and supervised data from SUNRGB-D, notably improves performance over our supervised baseline for the target scene. This is true for all of our experimental scenes in ScanNet in regions where our supervised baseline already provides a certain level of generalization for some of the viewpoints. For ``Scan 1'' we measure a significant increase in performance. To further demonstrate different use cases of S4-Net, we show that the network fine-tuned for ``Scan 1'' predicts high-grade segmentations also for ``Scan 2'' of the scene. 
The two scans are recorded independently which results in different camera paths for the recordings. As there are no direct neighboring frames between the two independently recorded scans, the results on  ``Scan 2'' demonstrate the ability of the S4-Net trained network to generalize to independently recorded scans of the same scene.



In our quantitative evaluations in Table~\ref{fig:scannet_quant}, we present results averaged over all of our experimental scenes. We observe that the S4-Net approach clearly overcomes the dataset bias correlated with the supervised approach as it demonstrates superior performance on the target scene in comparison to its supervised baseline. Not only does the performance increase for ``Scan 1'' but we also observe that the increase in performance for the images of ``Scan 2'' is as significant. Furthermore, by observing performance for different network architectures, we show that S4-Net can be applied to arbitrary segmentation network architectures.

We further demonstrate the benefits of our approach in Fig.~\ref{fig:scannet_qual} where we show some  qualitative results. We observe that, in areas where the supervised approach predicts very noisy predictions, our approach predicts consistent segmentations. This is the indicator that confident segmentation predictions are propagated to less confident viewpoints, and not the other way around.

Our experiments on ScanNet demonstrate that S4-Net is useful for different practical applications. First, the evaluations on ``Scan 1'' show that the approach is applicable for the use case of automatically labelling indoor scenes. The second application is that, once the network has converged for the target sequence, we can reliably segment new images of the scene without the need for the depth data.


\subsection{Evaluation on Additional Scene}

So far we have demonstrated that S4-Net works well for the chosen scenes from the ScanNet dataset. To show that S4-Net generalizes well, we also evaluate it on our own data. For this purpose, we captured and registered a living room area using an Intel\textsuperscript{\textregistered}RealSense\textsuperscript{TM} D400 series depth camera~\footnote{\href{https://software.intel.com/en-us/realsense/d400}{https://software.intel.com/en-us/realsense/d400}} and registered the scene using an implementation of a scene reconstruction pipeline~\cite{choi2015robust} from Open3D library~\cite{Zhou2018}. We refer to this scene as the ``Room'' dataset. 

\textbf{Data Split}. In line with the ScanNet experiments, we scanned the ``Room'' scene twice. ``Scan 1'' contains roughly $6000$ training images. For evaluation purposes, we then sampled $20$ images from ``Scan 1'' that capture different viewpoints of the scene, and we manually annotated them using the LabelMe annotation tool~\cite{russell2008labelme}. We annotated $20$ additional images from ``Scan 2'' that was recorded independently of ``Scan 1'' to further demonstrate the aspect of generalization across the scene. 

\begin{figure}[t]
    \centering
     \begin{subfigure}[b]{1.\linewidth}
     \centering
     \begin{minipage}{0.04\linewidth}
     \rotatebox[origin=t]{90}{Sup. baseline}
    \end{minipage}
    \begin{minipage}{0.31\linewidth}
    \includegraphics[width=\linewidth]{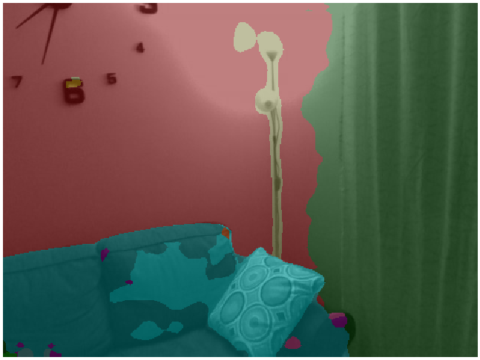}
    \end{minipage}
    \begin{minipage}{0.31\linewidth}
    \includegraphics[width=\linewidth]{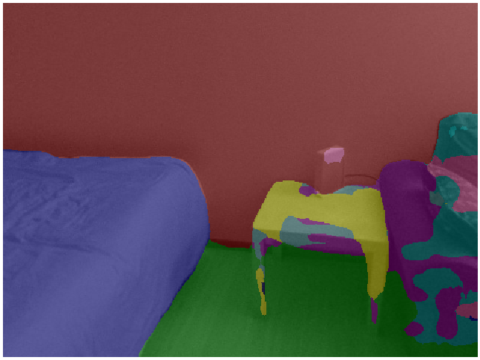}
    \end{minipage}
    \begin{minipage}{0.31\linewidth}
    \includegraphics[width=\linewidth]{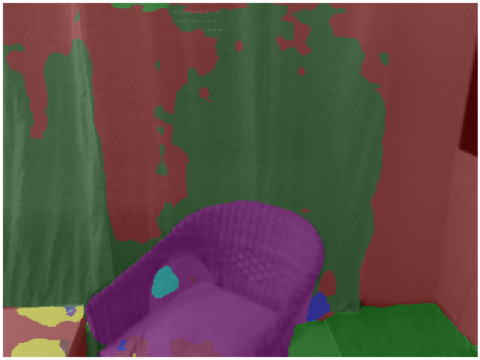}
    \end{minipage}
    \end{subfigure}
    
     \begin{subfigure}[b]{1.\linewidth}
     \centering
     \begin{minipage}{0.04\linewidth}
          \rotatebox[origin=t]{90}{\textit{S4-Net}}
    \end{minipage}
    \begin{minipage}{0.31\linewidth}
    \includegraphics[width=\linewidth]{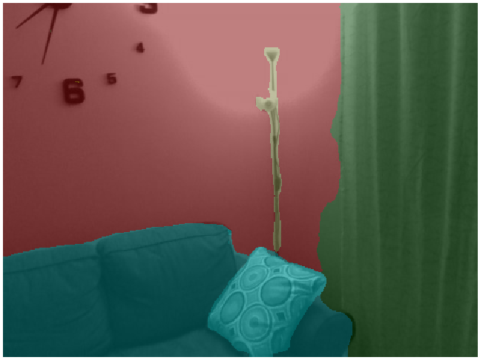}
    \end{minipage}
    \begin{minipage}{0.31\linewidth}
    \includegraphics[width=\linewidth]{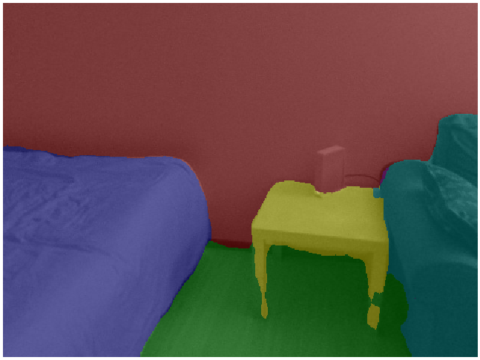}
    \end{minipage}
    \begin{minipage}{0.31\linewidth}
    \includegraphics[width=\linewidth]{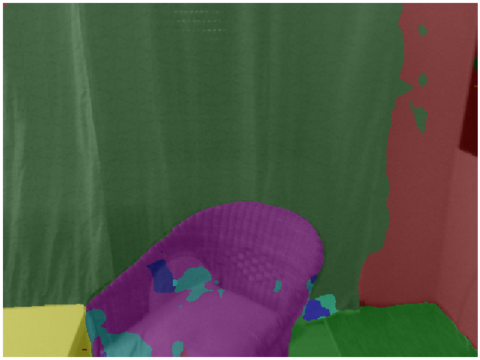}
    \end{minipage}
    \end{subfigure}
    
    \begin{subfigure}[b]{1.\linewidth}
     \centering
    \begin{minipage}{0.04\linewidth}
        \rotatebox[origin=t]{90}{\hspace{0.5cm} Man. annot.}
    \end{minipage}
    \begin{minipage}{0.31\linewidth}
    \includegraphics[width=\linewidth]{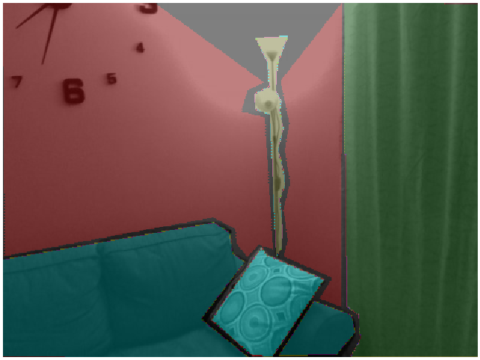}    \captionof{figure}{}
    \end{minipage}
    \begin{minipage}{0.31\linewidth}
    \includegraphics[width=\linewidth]{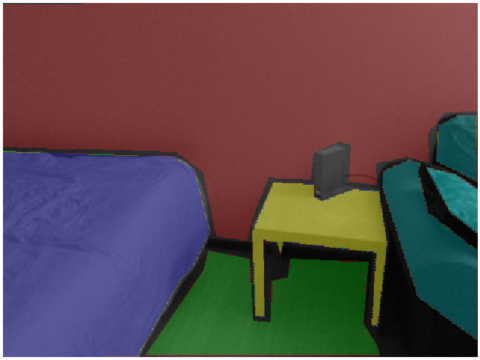}    \captionof{figure}{}
    \end{minipage}
    \begin{minipage}{0.31\linewidth}
    \includegraphics[width=\linewidth]{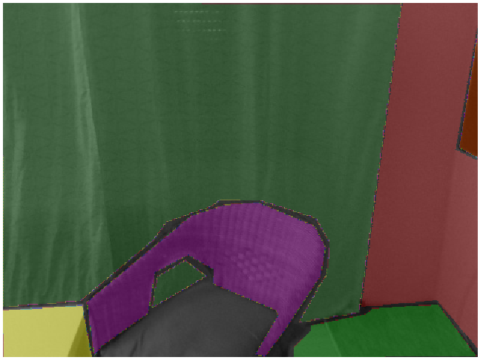}    \captionof{figure}{}
    \end{minipage}
    \end{subfigure}

    \caption{Qualitative evaluation on unseen images from ``Scan 2'' of the ``Room'' scene for the DeepLabV3+ network architecture. The supervised baseline already predicts high quality segmentations across the scene. However, the supervised baseline still predicts noisy or incorrect segmentations for some view points, especially due to partial visibility of objects, for example the sofa in (b) and the table in (c). S4-Net demonstrates notable improvements in these regions.}
    \label{fig:custom_qual}
\end{figure}

Table~\ref{fig:custom_quant} gives the results of our quantitative evaluation. We again observe a significant increase in performance over the supervised baseline approach for images of ``Scan 1'' and ``Scan 2''. Our qualitative evaluations in Fig.~\ref{fig:custom_qual} show many overall improvements. Even though our supervised baseline might pedict quality segmentations for specific viewpoints, for other viewpoints it fails completely as these data samples are not presented well throughout the SUNRGB-D dataset. In contrast, the S4-Net approach preserves quality segmentations in such regions. This further proves that the usage of geometric constraints is, indeed, a very powerful method for transfering knowledge from the supervised baseline to a new scene.

\begin{table*}[t]
 \centering
\begin{tabular}{@{}lcccccccrcccccc@{}}
  \toprule
  & \phantom{}&  \multicolumn{6}{c}{{\bf ``Scan 1'' (``Room'')}} & \phantom{a}& \multicolumn{6}{c}{{\bf ``Scan 2'' (``Room'')}}\\
  
    & \phantom{}&  \multicolumn{6}{c}{{(Unlabeled images during training)}} & \phantom{a}& \multicolumn{6}{c}{{(Excluded from training)}}\\
  
  && \sm{pix\_acc} & \sm{mean\_acc} & \sm{mIOU} & \sm{fwIOU} &&&&  \sm{pix\_acc} & \sm{mean\_acc} & \sm{mIOU} & \sm{fwIOU} &&   \\
  \midrule
    \multicolumn{6}{l}{\textbf{DeepLabV3+ network architecture}} &&&&&&&& \\ 
  Supervised baseline && $0.89$ & $0.757$ & $0.699$ & $0.827$ &&&& $0.817$ & $0.726$ & $0.66$ & $0.76$ && \\

  \textit{S4-Net}     && $\textbf{0.934}$ & $\textbf{0.846}$ & $\textbf{0.799}$ & $\textbf{0.884}$ &&&& $\textbf{0.938}$ & $\textbf{0.75}$ & $\textbf{0.719}$ & $\textbf{0.906}$ && \\
  
  \textit{S4-Net with depth pred.} && $0.922$ & $0.801$ & $0.757$ & $0.868$ &&&& $0.911$ & $0.755$ & $0.712$ & $0.867$ && \\
  
  \multicolumn{6}{l}{\textbf{PSPNet network architecture}} &&&&&&&& \\ 
  Supervised baseline && $0.862$ & $0.673$ & $0.597$ & $0.788$ &&&& $0.728$ & $0.629$ & $0.497$ & $0.651$ && \\

  \textit{S4-Net} && $\bf 0.888$ & $\bf 0.723$ & $\bf 0.645$ & $\bf 0.817$ &&&& $\bf 0.847$ & $\bf 0.708$ & $\bf 0.604$ & $\bf 0.782$ && \\

  \bottomrule
\end{tabular}

\caption{Quantitative evaluation on the ``Room'' scene. Similarly to our experiments on ScanNet, S4-Net demonstrates significant performance increase for both the images from ``Scan 1'' and ``Scan 2'' of the scene. We observe improvements for different network architectures and also when using S4-Net with depth prediction network.}
\label{fig:custom_quant}

\end{table*}

\subsection{Evaluation of S4-Net with Depth Prediction}

Furthermore, we evaluate the aspect of using depth predictions  for enforcing geometric constraints. As SUNRGB-D also contains depth ground truth data, it provides supervision for both the depth network and the segmentation network in this scenario. When enforcing geometric constraints for the target scenes, warping between different view points is performed by using depth predictions instead of the ground truth depth images. For this experiment, we found empirically that setting $\lambda_D$ to $0.1$ achieved satisfying quality for segmentation and depth predictions. In case of PSPNet, due to low accuracy of initial depth predictions, we excluded this part in our evaluations.

\begin{figure}[t]
    \centering
    \begin{subfigure}[b]{1.\linewidth}
     \centering
     \begin{minipage}{0.04\linewidth}
          \rotatebox[origin=t]{90}{Sup. baseline}
    \end{minipage}
    \begin{minipage}{0.31\linewidth}
    \includegraphics[width=\linewidth]{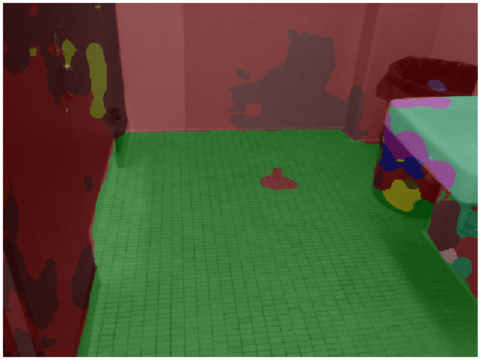}
    \end{minipage}
    \begin{minipage}{0.31\linewidth}
    \includegraphics[width=\linewidth]{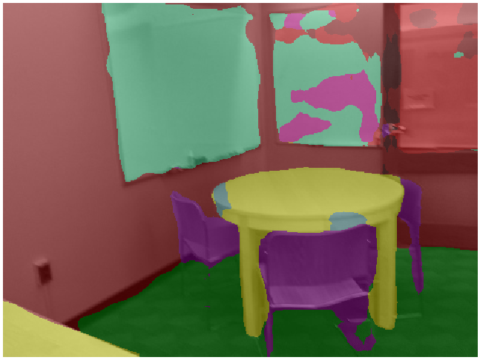}
    \end{minipage}
    \begin{minipage}{0.31\linewidth}
    \includegraphics[width=\linewidth]{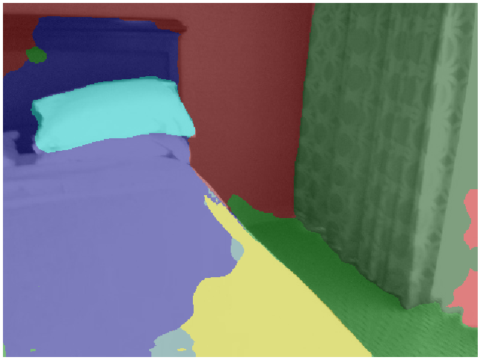}
    \end{minipage}
    \end{subfigure}
    
    \begin{subfigure}[b]{1.\linewidth}
     \centering
     \begin{minipage}{0.04\linewidth}
               \rotatebox[origin=t]{90}{\textit{S4-Net}}
    \end{minipage}
    \begin{minipage}{0.31\linewidth}
    \includegraphics[width=\linewidth]{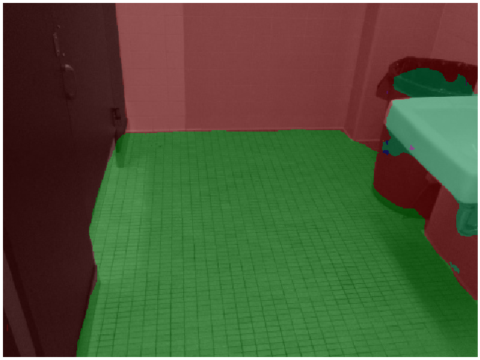}
    \end{minipage}
    \begin{minipage}{0.31\linewidth}
    \includegraphics[width=\linewidth]{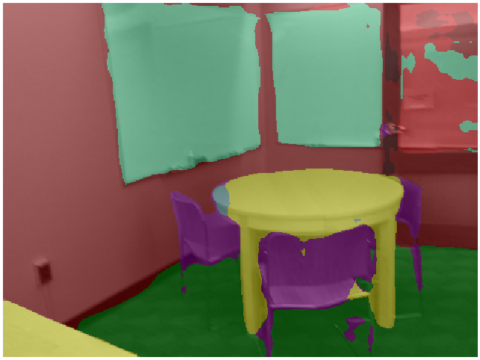}
    \end{minipage}
    \begin{minipage}{0.31\linewidth}
    \includegraphics[width=\linewidth]{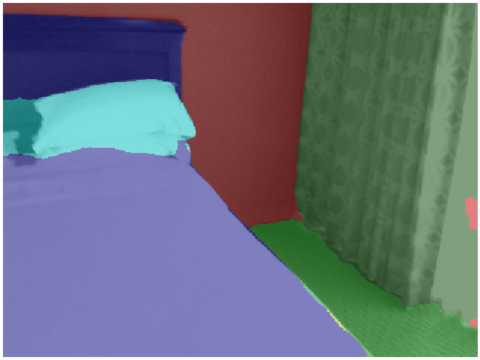}
    \end{minipage}
    \end{subfigure}
    
    \begin{subfigure}[b]{1.\linewidth}
     \centering
    \begin{minipage}{0.04\linewidth}
            \rotatebox[origin=t]{90}{\hspace{0.5cm} Man. annot.}
    \end{minipage}
    \begin{minipage}{0.31\linewidth}
    \includegraphics[width=\linewidth]{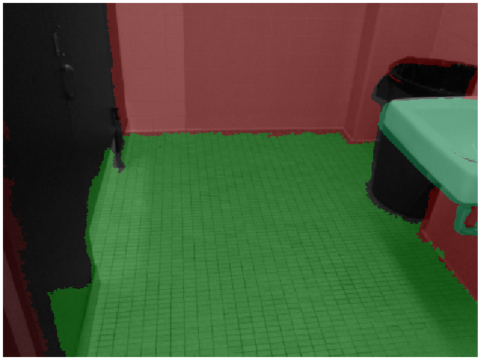}    
    \end{minipage}
    \begin{minipage}{0.31\linewidth}
    \includegraphics[width=\linewidth]{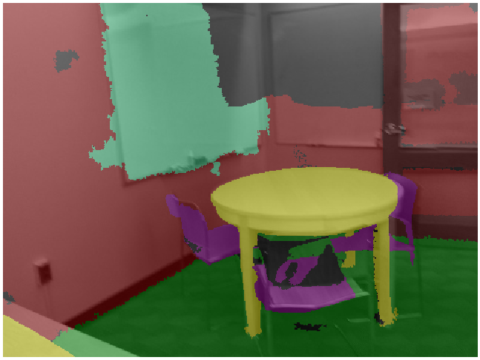}    
    \end{minipage}
    \begin{minipage}{0.31\linewidth}
    \includegraphics[width=\linewidth]{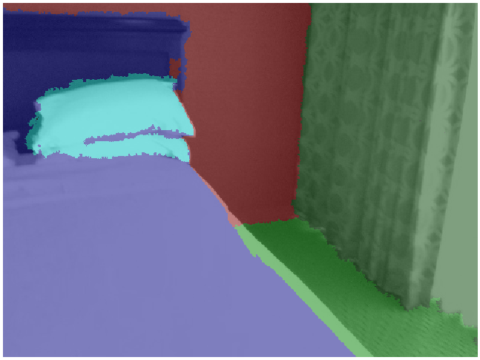}    
    \end{minipage}
    \end{subfigure}
    
    \caption{Qualitative results of S4-Net with depth predictions on  unseen images from ``Scan 2'' of the target scenes from ScanNet. Similarly to our previous observations, S4-Net predicts quality segmentations in many regions which are noisy, wrongly labeled or unlabeled in the manual annotations. }
    \label{fig:depth_pred}
\end{figure}

Our quantitative evaluations in Table~\ref{fig:scannet_quant} demonstrate comparable performance to S4-Net with depth ground truth for the ScanNet scenes. Similarly, in Table~\ref{fig:custom_quant} we observe that S4-Net with depth predictions shows significant improvements for the ``Room'' scene in comparison to the supervised baseline. In our qualitative results in Fig. ~\ref{fig:depth_pred} we visualize the results of S4-Net with depth predictions for the ScanNet scenes. Even though one would expect that noisy depth predictions considerably decrease the quality of geometric constraints, S4-Net still demonstrates quality improvements in this scenario that is comparable to our results when using ground truth depth for enforcing geometric constraints. Even though initial depth predictions for the supervised baseline are noisy, S4-Net also learns better depth predictions for the target scene. Hence, geometric constraints on semantic segmentation improve during training enabling convergence for S4-Net. For unseen ``Scan 2'' sequences from ScanNet, the Root Mean Square~(RMS) error drops from $0.61$ to $0.4$ on average after applying S4-Net. For ``Scan 2'' images from ``Room'' scene, the average RMS error drops from $0.58$ to $0.49$. We show further quantitative and qualitative results on depth predictions in supplementary material.


\section{Conclusion}

We showed that semi-supervised learning is a good theoretical framework for enforcing geometric constraints and for adapting semantic segmentation to new scenes. We also investigated a potential problem which could appear with such semi-supervised constraints on non-annotated sequences. It would be possible that the learning may assign labels which are consistent among views, but wrong. Our experiments have shown that this is only very rarely the case. Instead, the semi-supervised contraints yield significant improvements, without the need for additional manual labels. This is possible because the network can learn to propagate labels from locations where it is confident to more difficult locations. 

In summary, our S4-Net approach yields quality labels across given target sequences which makes it very interesting for the task of sequence labelling. The segmentation network trained with S4-Net also generalizes nicely to unseen images of the target scene. This makes our approach useful for applications relying on semantic segmentation, for example in robotics and augmented reality. Finally, we have shown that the attractive idea of enforcing geometric constraints by means of depth predictions produces satisfying segmentations and achieves accuracy that is comparable to the accuracy when using ground truth depth information.

\section*{Acknowledgment}
This work was supported by the Christian Doppler Laboratory for Semantic 3D Computer Vision, funded in part by Qualcomm Inc.

\newpage

{\small
\bibliographystyle{ieee}
\bibliography{egbib}
}

\clearpage


\section*{Supplementary material}

# Casting Geometric Constraints in Semantic Segmentation as Semi-Supervised Learning - Supplementary Material

Sinisa Stekovic[1]   Friedrich Fraundorfer[1]   Vincent Lepetit[2,1]

[1]Institute for Computer Graphics and Vision, Graz University of Technology, Graz, Austria
[2]Université Paris-Est, École des Ponts ParisTech, Paris, France

{sinisa.stekovic, fraundorfer, lepetit}@icg.tugraz.at

## 1. Temporal Constraints for Semantic Segmentation

As shown in Figure 1, even though comparing predictions of noisy representations of input images has been shown as powerful temporal constraints in semi-supervised learning for image classification problems [9, 11, 1], such constraints are dangerous for semantic segmentation as they induce significant prediction error on pixelwise level. In comparison, our geometric constraints show notably reduced prediction noise which significantly improves probability of convergence for S4-Net.

## 2. Network Initialization And Training

**Initialization.** To initialize the network $f(.;\Theta)$, we first train it only based on the $L_S$ loss term. This avoids the problem of converging to a bad local minimum introduced by the term $L_G$. As it is the case with other consistency-based models, minimizing $L_G$ may fall in a solution where a single class is predicted for all the image locations. Even though tuning hyper-parameter $\lambda$ more carefully might resolve this problem, we noticed that using this pre-training step makes the convergence to a correct model much easier.

When predicting depth maps, the encoder is shared between the depth network and the segmentation network. The depth decoder has the same architecture as the segmentation decoder, but they do not share any parameters. To initialize the networks $f(.;\Theta)$ and $f_d(.;\Theta_d)$, we train them using only the supervised loss term $L_S + L_{DS}$. The full loss term $L + L_D$ is utilized afterwards.

**Training Details.** At every iteration, we sample a batch of 4 examples from each of the involved datasets. The input images are resized to $480 \times 360$. For better convergence, we pre-train the network using only the $L_S$ term on SUNRGB-D training set until convergence on the SUNRGB-D test set. We use the Adam optimizer [8] with an initial learning rate of $10^{-4}$ and momentum of $0.9$ for this step. In our experiments, we refer to this network as the supervised

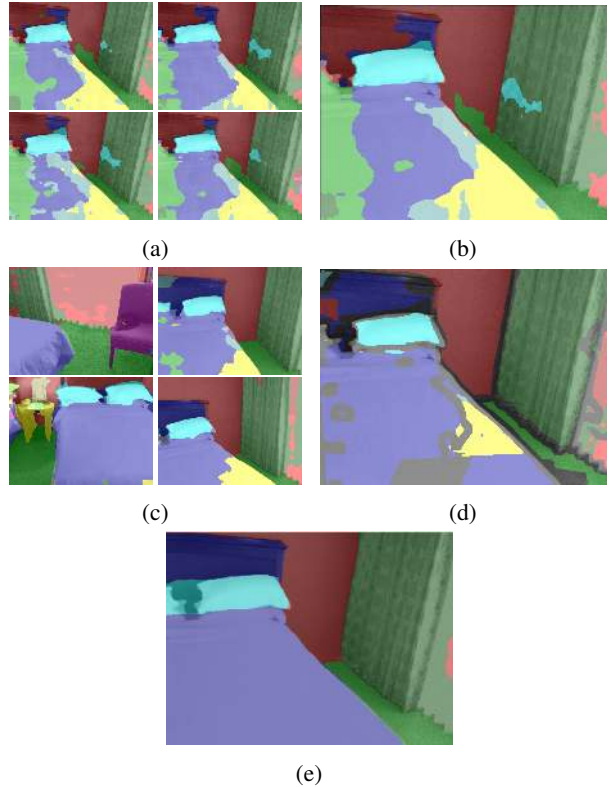

(a)  (b)

(c)  (d)

(e)

Figure 1: Semi-supervised terms for semantic segmentation. (a) Simply comparing predictions of noisy representations of input training images does not exploit the geometric constraints of the problem. (b) Merging predictions of noise-added inputs induces further errors in pixelwise segmentation and does not perform well as a temporal constraint. (c) and (d) Merging information from different views is more useful as a temporal constraint and enables convergence for S4-Net as shown in (e).

baseline. When learning to segment a target scene with our semi-supervised approach, we fine-tune the network with

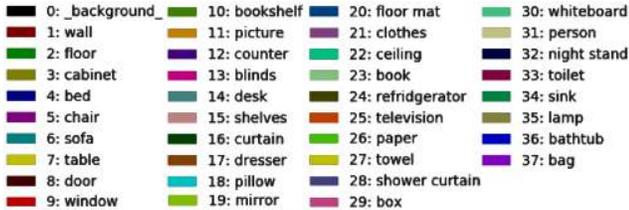

Figure 2: Colormap used for visualizing segmentations. _background_ is not one of the categories but it is used to visualize regions of images without manual annotations.

initial learning rate of $10^{-5}$ until no further improvements in performance are notable for the target scene. We found that setting parameter $\lambda$ to 0.01 balances the loss terms during this stage of training. For the geometric consistency loss term $L_G$, for a given sample $e$, we randomly sample 4 neighbouring viewpoints with a minimum overlapping region of 25% relative to the viewpoint of $e$. Through this, we achieve a trade-off between stability of the training process and the computational costs. In the case of PSPNet, all loss terms are applied to the auxiliary prediction branch as well. According to [14], this step improves gradient back-propagation during training. Furthermore, when applying S4-Net with PSPNet, due to very high computational costs, we set the batch size for the dataset $\mathcal{U}$ to 3.

## 3. Evaluation Metrics

For quantitative evaluation, we report common evaluation metrics. For a given image, we compare segmentation prediction to the corresponding ground truth annotation using following metrics:

- Pixel accuracy: $pix\_acc = \sum_i \frac{n_{ii}}{t_i}$ ;
- Mean accuracy: $mean\_acc = \frac{1}{n_c} \sum_i \frac{n_{ii}}{t_i}$ ;
- Mean IOU: $mIOU = \frac{1}{n_c} \sum_i \frac{n_{ii}}{t_i + \sum_j n_{ji} - n_{ii}}$ ;
- Frequency weighted IOU:
  $fwIOU = \frac{1}{\sum_i t_i} \sum_i \frac{t_i n_{ii}}{t_i + \sum_j n_{ji} - n_{ii}}$ ,

where $n_{ij}$ is the number of pixels of class $i$ predicted as class $j$, $n_c$ the number of classes, and $t_i$ the number of pixels that belong to class $i$. Results are then averaged accross the given dataset to measure the performance.

## 4. Failure Cases

As demonstrated in Figure 3, we observe that S4-Net fails when the estimated camera poses are inaccurate for the registered scene. Hence, the performance of S4-Net is tightly related to the performance of underlying scene registration pipeline.

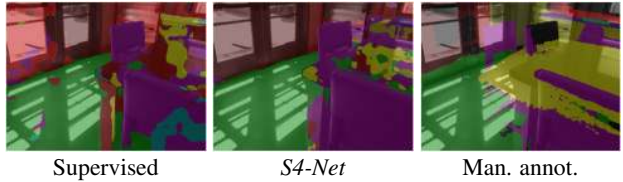

Supervised        S4-Net        Man. annot.

Figure 3: Failure case of S4-Net due to noisy geometric constraints caused by high level of reflectivity in scene0011. Even the manual annotations are inaccurate which indicates high error for camera poses across the scene.

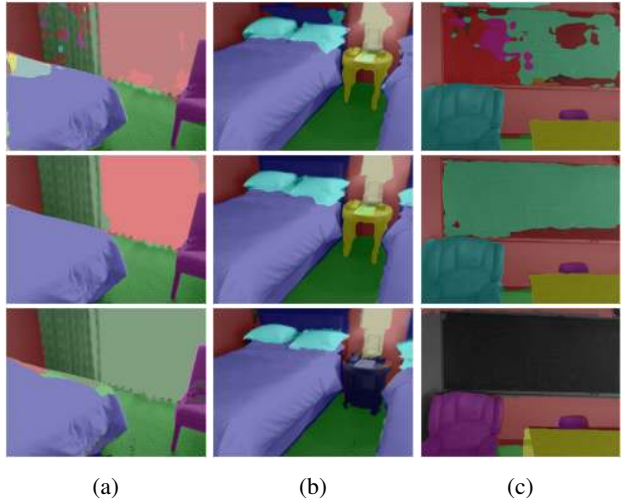

(a)          (b)          (c)

Figure 4: Even though comparisons with ground truth annotations might indicate wrong segmentation predictions, our segmentation predictions are often visually very appealing. In (a), S4-Net wrongly predicts window that is actually occluded by a transparent curtain. In (b), both the supervised baseline and S4-Net confuse the night stand with a table. In (c), S4-Net predicts a sofa where the manual annotation suggests a chair and the actual definition of the category for this object would be a 'single seater sofa'. As this category is not one of the given categories for the task, we find both of these alternatives equally correct.

Furthermore, the categories colormap in Figure 2 clearly reveals many object categories that might be interchangeable with each other. We show in Figure 4 that S4-Net sometimes predicts these 'alternative' categories instead of the ones proposed in manual annotations.

## 5. Quantitative Results for Individual Scenes from ScanNet

Table 1 presents details on all of the ScanNet scans that were used for the experiments. Table 2 presents quantitative experiments on the individual scenes from ScanNet. We observe that S4-Net consistently outperforms its supervised baseline for all of the scenes.

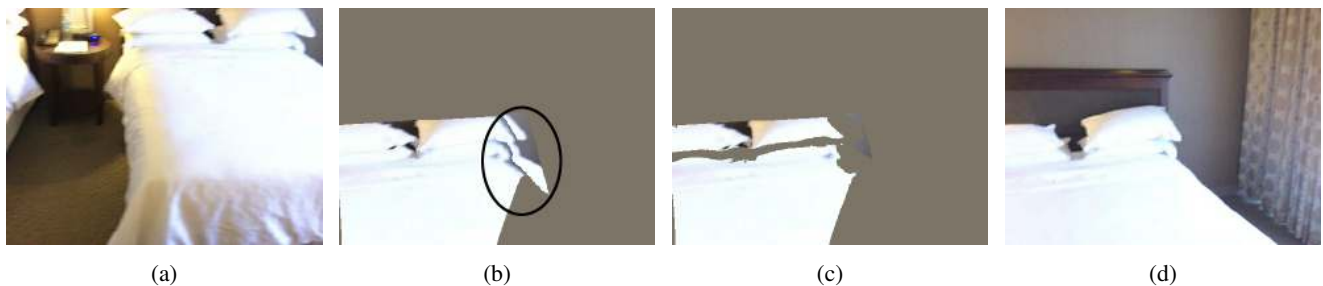

| (a) | (b) | (c) | (d) |

Figure 5: When warping the source view in (a) to the target view (d), we notice that occlusions can lead to association of incorrect correspondences between two views as encircled in (b). When learning to predict depth with $L_{INT}$ term, this results in errors caused by intensity differences of incorrect correspondences. We resolve this by masking out correspondences with inconsistent segmentation predictions as shown in (c).

| ScanNet Scene | "Scan 1" #Samples | "Scan 2" #Samples |
|---|---|---|
| scene0000 | 5578 | 5920 |
| scene0006 | 2161 | 2155 |
| scene0009 | 980 | 920 |
| scene0011 | 2374 | 2759 |
| scene0022 | 1896 | 1339 |
| scene0030 | 2498 | 1648 |

Table 1: Number of samples for the utilized scans from ScanNet.

## 6. Segmentation Mask for Learning Depth Prediction

We observe that the $L_{INT}$ loss term is very sensitive to occlusions in the scene. Even though existing works utilizing this term for monocular depth estimation achieve good results [15, 5, 13, 10, 12, 6], the baseline between the two camera views is very small. This reduces the negative influence of occlusions between the views and, hence, the network still learns quality depth estimations. As the baseline between the two views increases, occlusions in the scene might have more negative influence on the geometric constraints as we show in Figure 5b. Hence, in order to deal with occlusions in the image, we apply the $L_{INT}$ term only to the target image pixels where predicted segmentations are consistent with each other and further away from segmentation borders. In Figure 5c we show that such mask successfully masks out the affected regions of the warped image.

## 7. Evaluation of Depth Predictions

In Table 3 and Figure 6 we demonstrate that fine-tuning S4-Net with depth predictions for the target scene also improves depth estimations across the scene. Although applying a smoothness loss term would further improve our depth estimations [4, 15, 5, 13, 10, 12, 6], the quality of the learned predictions is good enough for enforcing geometric constraints on semantic segmentation.

|  | "Scan 1" (ScanNet) | | | | "Scan 2" (ScanNet) | | | |
| --- | --- | --- | --- | --- | --- | --- | --- | --- |
|  | pix_acc | mean_acc | mIOU | fwIOU | pix_acc | mean_acc | mIOU | fwIOU |
| **DeepLabV3+ network architecture** | | | | | | | | |
| **Supervised baseline** | | | | | | | | |
| scene0000 (Apartment) | 0.67 | 0.505 | 0.398 | 0.591 | 0.709 | 0.536 | 0.435 | 0.643 |
| scene0006 (Hotel room) | 0.751 | 0.629 | 0.54 | 0.683 | 0.723 | 0.617 | 0.511 | 0.653 |
| scene0009 (Bathroom) | 0.896 | 0.811 | 0.73 | 0.855 | 0.908 | 0.851 | 0.75 | 0.86 |
| scene0011 (Kitchen) | 0.675 | 0.532 | 0.407 | 0.575 | 0.716 | 0.555 | 0.442 | 0.622 |
| scene0022 (Lounge) | 0.872 | 0.751 | 0.657 | 0.81 | 0.829 | 0.697 | 0.585 | 0.74 |
| scene0030 (Study room) | 0.726 | 0.574 | 0.467 | 0.638 | 0.744 | 0.647 | 0.539 | 0.662 |
| *Average* | 0.765 | 0.634 | 0.533 | 0.692 | 0.772 | 0.651 | 0.544 | 0.697 |
| ***S4-Net*** | | | | | | | | |
| scene0000 (Apartment) | 0.726 | 0.565 | 0.465 | 0.642 | 0.752 | 0.579 | 0.486 | 0.68 |
| scene0006 (Hotel room) | 0.816 | 0.725 | 0.647 | 0.767 | 0.782 | 0.679 | 0.599 | 0.73 |
| scene0009 (Bathroom) | 0.907 | 0.88 | 0.785 | 0.873 | 0.927 | 0.909 | 0.813 | 0.889 |
| scene0011 (Kitchen) | 0.701 | 0.553 | 0.431 | 0.595 | 0.741 | 0.582 | 0.47 | 0.643 |
| scene0022 (Lounge) | 0.896 | 0.805 | 0.708 | 0.833 | 0.831 | 0.724 | 0.608 | 0.745 |
| scene0030 (Study room) | 0.77 | 0.596 | 0.505 | 0.689 | 0.787 | 0.674 | 0.58 | 0.708 |
| *Average* | **0.803** | **0.687** | **0.59** | **0.733** | **0.803** | **0.691** | **0.593** | **0.732** |
| ***S4-Net with Depth Predictions*** | | | | | | | | |
| scene0000 (Apartment) | 0.713 | 0.538 | 0.438 | 0.63 | 0.746 | 0.563 | 0.471 | 0.677 |
| scene0006 (Hotel room) | 0.803 | 0.715 | 0.636 | 0.754 | 0.779 | 0.674 | 0.595 | 0.728 |
| scene0009 (Bathroom) | 0.932 | 0.89 | 0.804 | 0.902 | 0.956 | 0.924 | 0.848 | 0.925 |
| scene0011 (Kitchen) | 0.659 | 0.545 | 0.414 | 0.542 | 0.699 | 0.564 | 0.443 | 0.585 |
| scene0022 (Lounge) | 0.9 | 0.791 | 0.701 | 0.842 | 0.825 | 0.704 | 0.59 | 0.738 |
| scene0030 (Study room) | 0.759 | 0.593 | 0.493 | 0.674 | 0.775 | 0.674 | 0.569 | 0.695 |
| *Average* | 0.794 | 0.679 | 0.581 | 0.724 | 0.797 | 0.684 | 0.586 | 0.725 |
| **PSPNet network architecture** | | | | | | | | |
| **Supervised baseline** | | | | | | | | |
| scene0000 (Apartment) | 0.618 | 0.453 | 0.346 | 0.534 | 0.664 | 0.486 | 0.38 | 0.592 |
| scene0006 (Hotel room) | 0.683 | 0.586 | 0.472 | 0.6 | 0.67 | 0.58 | 0.458 | 0.588 |
| scene0009 (Bathroom) | 0.888 | 0.81 | 0.705 | 0.832 | 0.904 | 0.814 | 0.725 | 0.847 |
| scene0011 (Kitchen) | 0.635 | 0.488 | 0.362 | 0.531 | 0.66 | 0.498 | 0.386 | 0.565 |
| scene0022 (Lounge) | 0.849 | 0.724 | 0.618 | 0.777 | 0.821 | 0.685 | 0.569 | 0.726 |
| scene0030 (Study room) | 0.686 | 0.522 | 0.414 | 0.591 | 0.703 | 0.599 | 0.478 | 0.609 |
| *Average* | 0.727 | 0.597 | 0.486 | 0.644 | 0.737 | 0.61 | 0.499 | 0.654 |
| ***S4-Net*** | | | | | | | | |
| scene0000 (Apartment) | 0.716 | 0.523 | 0.42 | 0.63 | 0.731 | 0.541 | 0.445 | 0.66 |
| scene0006 (Hotel room) | 0.76 | 0.679 | 0.565 | 0.686 | 0.722 | 0.629 | 0.511 | 0.644 |
| scene0009 (Bathroom) | 0.91 | 0.872 | 0.748 | 0.863 | 0.934 | 0.879 | 0.792 | 0.888 |
| scene0011 (Kitchen) | 0.68 | 0.523 | 0.4 | 0.567 | 0.722 | 0.552 | 0.44 | 0.617 |
| scene0022 (Lounge) | 0.897 | 0.767 | 0.668 | 0.832 | 0.817 | 0.694 | 0.569 | 0.726 |
| scene0030 (Study room) | 0.724 | 0.526 | 0.433 | 0.627 | 0.752 | 0.617 | 0.519 | 0.66 |
| *Average* | **0.781** | **0.648** | **0.539** | **0.701** | **0.78** | **0.652** | **0.546** | **0.699** |

Table 2: Quantitative evaluation on the target scenes from ScanNet. The table shows results calculated per individual scene and average performance over all of the target scenes. The numbers for individual scenes indicate that S4-Net consistently outperforms the supervised baseline for all of the scenes.

|  | "Scan 2" (ScanNet) | | | | | | |
|---|---|---|---|---|---|---|---|
|  | Lower is better | | | | Higher is better | | |
|  | Abs Rel | Sq Rel | RMSE | RMSE log | $\delta < 1.25$ | $\delta < 1.25^2$ | $\delta < 1.25^3$ |
| **DeepLabV3+ network architecture** | | | | | | | |
| **Supervised baseline** | | | | | | | |
| scene0000 (Apartment) | 0.296 | 0.192 | 0.524 | 0.271 | 0.449 | 0.918 | 0.99 |
| scene0006 (Hotel room) | 0.358 | 0.28 | 0.609 | 0.321 | 0.374 | 0.857 | 0.962 |
| scene0009 (Bathroom) | 0.329 | 0.204 | 0.532 | 0.293 | 0.257 | 0.958 | 0.993 |
| scene0011 (Kitchen) | 0.319 | 0.322 | 0.711 | 0.303 | 0.444 | 0.9 | 0.967 |
| scene0022 (Lounge) | 0.324 | 0.37 | 0.723 | 0.297 | 0.441 | 0.932 | 0.975 |
| scene0030 (Study Room) | 0.298 | 0.187 | 0.574 | 0.274 | 0.357 | 0.963 | 0.994 |
| *Average* | 0.321 | 0.259 | 0.612 | 0.293 | 0.387 | 0.921 | 0.98 |
| ***S4-Net with Depth Predictions*** | | | | | | | |
| scene0000 (Apartment) | 0.115 | 0.048 | 0.269 | 0.131 | 0.885 | 0.988 | 0.998 |
| scene0006 (Hotel room) | 0.122 | 0.09 | 0.452 | 0.174 | 0.886 | 0.958 | 0.975 |
| scene0009 (Bathroom) | 0.121 | 0.053 | 0.271 | 0.133 | 0.923 | 0.993 | 0.996 |
| scene0011 (Kitchen) | 0.168 | 0.159 | 0.553 | 0.219 | 0.804 | 0.938 | 0.975 |
| scene0022 (Lounge) | 0.179 | 0.199 | 0.535 | 0.195 | 0.823 | 0.97 | 0.982 |
| scene0030 (Study Room) | 0.134 | 0.057 | 0.341 | 0.151 | 0.897 | 0.99 | 0.999 |
| *Average* | **0.14** | **0.101** | **0.404** | **0.167** | **0.87** | **0.973** | **0.987** |
|  | "Scan 2" ("Room") | | | | | | |
|  | Abs Rel | Sq Rel | RMSE | RMSE log | $\delta < 1.25$ | $\delta < 1.25^2$ | $\delta < 1.25^3$ |
| **DeepLabV3+ network architecture** | | | | | | | |
| Supervised baseline | 0.419 | 0.263 | 0.584 | 0.358 | 0.153 | 0.839 | 0.99 |
| *S4-Net with Depth Predictions* | **0.337** | **0.183** | **0.495** | **0.298** | **0.307** | **0.925** | **0.993** |

Table 3: Quantitative evaluation for depth predictions on the target scenes. Results indicate improvements in the depth domain for S4-Net with depth predictions. Hence, we obtain quality geometric constraints for learning semantic segmentation.

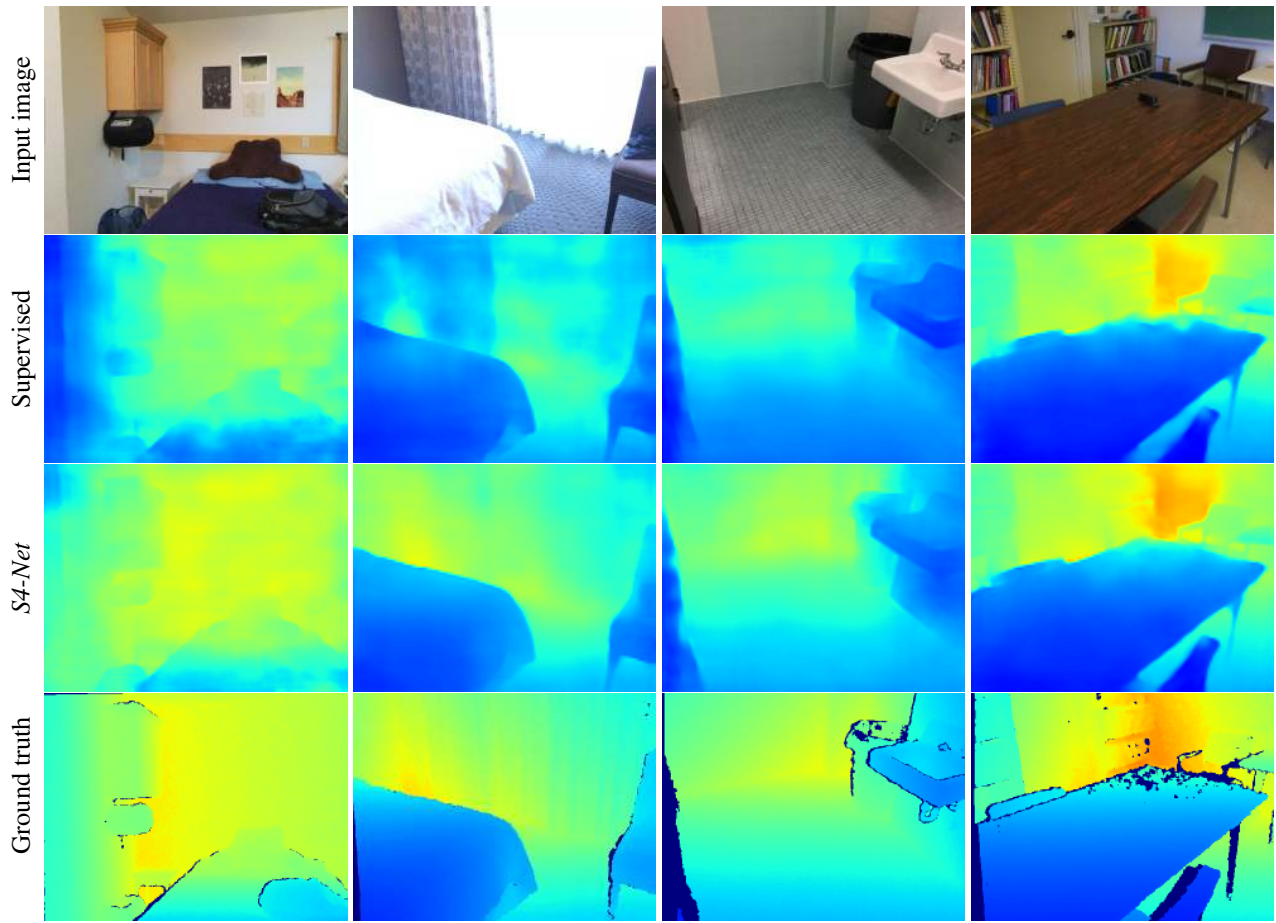

Figure 6: Qualitative evaluation of our depth predictions on unseen images of the target scenes from ScanNet for the DeepLabV3+ network architecture. After applying S4-Net with depth predictions to the target scenes, we observe improvements also in the depth domain.



\end{document}